# Transfer Learning-Based Crack Detection by Autonomous UAVs

F. Kucuksubasi[a] and A.G. Sorguc[b]

[a,b] Department of Architecture, Middle East Technical University, Turkey
E-mail: fatihk@metu.edu.tr, arzug@metu.edu.tr

**Abstract** – Unmanned Aerial Vehicles (UAVs) have recently shown great performance collecting visual data through autonomous exploration and mapping in building inspection. Yet, the number of studies is limited considering the post processing of the data and its integration with autonomous UAVs. These will enable huge steps onward into full automation of building inspection. In this regard, this work presents a decision making tool for revisiting tasks in visual building inspection by autonomous UAVs. The tool is an implementation of fine-tuning a pretrained Convolutional Neural Network (CNN) for surface crack detection. It offers an optional mechanism for task planning of revisiting pinpoint locations during inspection. It is integrated to a quadrotor UAV system that can autonomously navigate in GPS-denied environments. The UAV is equipped with onboard sensors and computers for autonomous localization, mapping and motion planning. The integrated system is tested through simulations and real-world experiments. The results show that the system achieves crack detection and autonomous navigation in GPS-denied environments for building inspection.

## 1   Introduction

Inspection of buildings throughout their lifecycle is vital in terms of human safety as the number of structures increases expeditiously. In line with this objective, periodic inspections are essential for residents' safety. For instance, systematic bridge inspections are done periodically in six years to detect structural cracks [1].

In this context, Unmanned Aerial Vehicles (UAVs) have been widely used in inspection operations in the last decade since their workspace is superior than that of ground vehicles. Today, UAVs are employed especially in visual building inspections by utilizing onboard cameras.

Moreover, the robotics community has increased the automated capabilities of the UAVs in terms of data acquisition and processing for inspection. In [1], a micro helicopter using computer vision approaches to be able to inspect bridges is presented. In [2], authors introduced a UAV system for inspecting culverts utilizing GPS, LIDAR and IMU. The data acquired from these sensors are fused to estimate the state enabling autonomous outdoor navigation.

A methodology to monitor the changes due to corrosion damages on industrial plants by using UAV is presented [3]. Images acquired at different instances are aligned through geometric transformation to highlight the changes above a threshold which is automatically determined by assuming damages that have usually different aspects with respect to the surrounding structures. In [4], researchers demonstrated a quadrotor MAV integrated with a stereo camera configuration that can explore GPS-denied indoor environments. They validated the system by autonomous flights inside an industrial boiler.

Recently, a lot of effort is put on crack detection using UAVs. A hybrid image processing technique that estimates crack width while decreasing the loss in crack length information is reported [5]. Another approach for crack detection and mapping using UAVs is presented in [6]. However, the previous works mainly focus on either autonomous navigation or post-processing of the acquired data (i.e. defect detection) by mostly using traditional image processing techniques (as in [7] & [8]) that may fail in different lighting conditions and/or materials.

In this context, the aim of this research is to achieve autonomous navigation and revisit motion planning of UAVs to surface crack locations in order to perform automated building inspection operations since the inspections are generally periodic and requires revisiting for close examination. This revisit task planning strategy enables UAVs to autonomously navigate in different environments while proposing a decision making tool by crack detection. The major contribution of this dissertation can be stated as an implementation of autonomous building inspection considering not only the data acquisition phase but also revisiting crack locations by transfer learning.

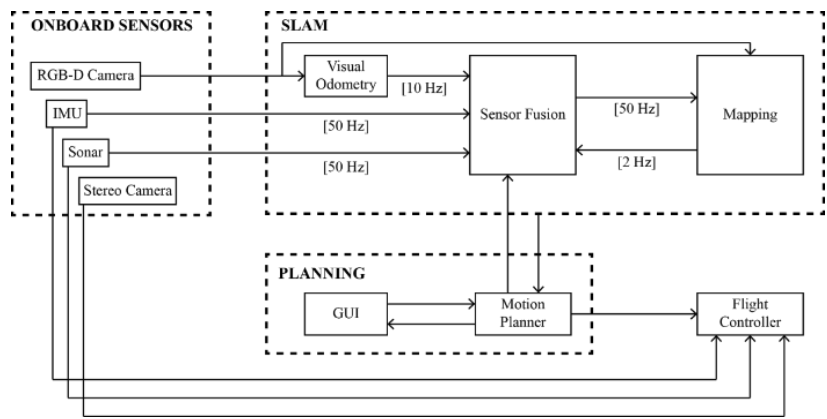

Figure 1. Schematic of the high-level system architecture

In this regard, an autonomously navigating quadrotor UAV is developed to be able to revisit pinpoint locations. For this purpose, SLAM using onboard visual-inertial localization and mapping to explore the environment in which the UAV is located and motion planning with obstacle avoidance are applied. Transfer learning approach is used to identify surface cracks from images so that possible revisiting locations can be determined for high-level decision making during inspections. Finally, a commercial quadrotor UAV is integrated with onboard sensors and computers in order to validate and verify the methods by testing.

## 2 System Overview

The necessary software for autonomous navigation and crack detection is both developed and implemented from open source libraries and packages supported by the community in the scope of this study. Figure 1 presents the schema of the software architecture for the overall system. all the computations are done onboard except from GUI and image classifier which runs on a ground station computer. The developed software components for this research is open-source and available online [9].

The flight controller of the quadrotor platform [10] ensures the low-level (attitude and velocity) control of the vehicle. Modified open-source software is implemented for SLAM and motion planning strategies. RGB-D camera is the source for the visual odometry and mapping processes, and it is fused with onboard IMU and ultrasonic sensor for state estimations by an Extended Kalman Filter. Also, a CNN is employed as the image classifier for surface crack detection. It presents an optional support mechanism for task planning of revisiting locations during inspection. It is built on top of autonomous navigation capability of the UAV with a user interface.

A graphical user interface that runs on a ground station computer is developed for high-level planning of the revisiting tasks. The interface wraps the capabilities of the system and enables users to utilize it without a prior knowledge of robotics. Functional callbacks for planning, motion control and other features for visualization purposes such as live video stream are other aspects derived from this interface.

The integrated system aims autonomous navigation with onboard computations. The objectives of autonomous navigation of the UAV in GPS-denied environments are determined as follows:

1. Real-time state estimation of the UAV during flight,
2. Mapping of the environment in which the UAV operates for global localization and motion planning,
3. Motion planning with obstacle avoidance to target locations of revisiting.

In order to achieve these objectives, software is developed and implemented on top of open source software packages and algorithms presented by the robotics community through Robot Operating System (ROS) so that the software can be modular and extendable. Implementation of the software throughout this work is achieved by using ROS framework.

Attitude control is achieved by using velocity control mode of the onboard flight controller in the employed UAV. A graph-based RTAB-MAP [11] ROS node is used both for visual odometry and mapping. A RGB-D camera is integrated with the UAV for visual odometry and onboard IMU is fused with visual odometry with an Extended Kalman Filter [12] at 50 Hz.

Motion planning is achieved with an implementation of MoveIt! [13] ROS node with OMPL [14] backend. A Universal Robot Description Format (URDF) of the UAV is constructed for collision checking. An algorithm is developed for interoperability of the motion planner and the revisit planner.

## 3 Revisit Planning

The proposed revisit planning workflow in Figure 3 is demonstrated step-by-step as follow:

1. Images acquired during flight are shown in the GUI for users to pick a revisiting location. Each image corresponds a location (position and orientation) in the algorithm since these images are previously linked with locations.
2. (Optional) A crack detector can step in to process images if a user aims to use the crack detection approach as a support for decision making of locations to revisit.
3. If Step-2 is fulfilled, the GUI visualizes the new cracks on a new set of images.
4. By the corresponding image, a location for revisiting is found. Then, the GUI delivers this goal to the motion planner.
5. The GUI receives an obstacle-free optimal trajectory as the form of waypoints if available.
6. For UAV to cover the path, the required velocities between the waypoints are calculated by an algorithm. In case of being successful or having no feasible motion plan, the GUI reports feedback.

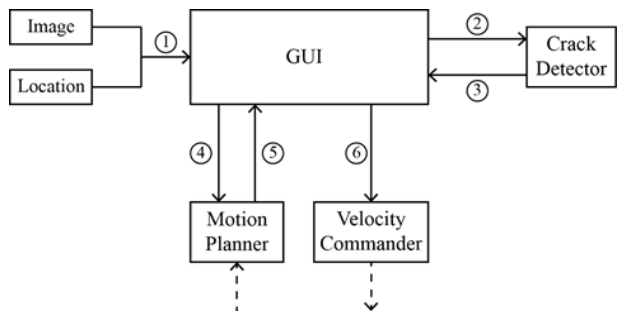

Figure 3. Workflow of the revisit planning strategy

For the step 1, Matching the images with their corresponding locations is essential for this strategy. A ROS node is developed in order to match images with poses (position and orientation) where the UAV had been visited during inspection for possible revisiting in future missions. This algorithm subscribes to both visual-inertial odometry and RGB images. Then, it matches them using Approximate Time Synchronizer message filter of ROS in a predefined period (2 Hz in this implementation). The images saved with their corresponding poses in favor of the revisit planner.

In order to provide the step 4&5, an algorithm is developed using Python interface of MoveIt!. It sends the goal to the motion planner and computes the velocities from the corresponding trajectory. It runs at the backend of the GUI.

## 4 Crack Detection

Crack detection is one of the most common objectives in building inspection. Increasing efficiency of visual inspection can be achieved by decreasing the time spent for post-processing on captured images during flight.

In this context, Convolutional Neural Networks (CNN) are one of the most commonly used architectures since they can overcome most of the contemporary challenges in crack detection [15]. They are getting more accurate and robust for image classification in recent years.

On the one hand, CNNs are easy to train and can be applied from open source libraries. On the other hand, training an entire CNN from scratch is not preferable for the majority because it is comparatively difficult to have a sufficient amount of data. Transfer Learning has eventually emerged. It uses a pretrained network on a large dataset (e.g. ImageNet which contains 1.2 million images with 1000 categories) as an initialization or a fixed feature extractor for the newly created network. Fine-tuning is one of the methods in Transfer Learning which is used as a complementary of CNN in this study. For working efficiently with small datasets and using a pretrained networks that assures time effectiveness, this approach becomes appropriate in the scope of this research.

### 4.1 Training the CNN

In this work, InceptionV3 [16] network model with ImageNet weights is fine-tuned since it has relatively high performance in top-1 validation accuracy than most of the top scoring single-model architectures (Figure 4). Keras with TensorFlow backend is used for the implementation. Graphics Processing Units are utilized in training sessions. For this purpose, a modified version of [17] is used.

In the fine-tuning of the CNN, 'Crack' and 'NonCrack' classes are designated for the image classifier. The training dataset is collected from Middle East Technical University campus buildings. It includes 582 images with cracks, and 458 images without cracks (Figure 5). Since a poor performance was observed on brick wall images in the first implementation, a group of brick wall images are added to the 'NonCrack' dataset in order to detect cracks on brick materials. Additionally, data augmentation function is applied to increase the number of the data for greater performance.

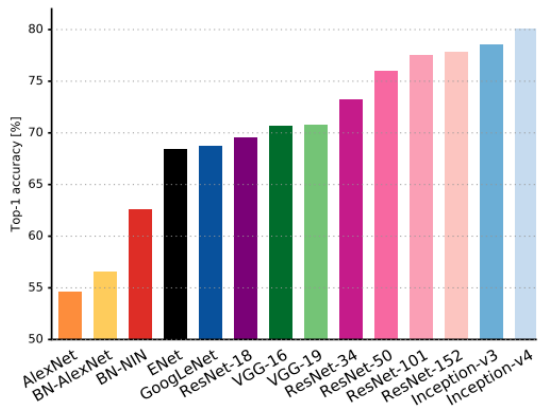

Figure 4. 13 Single-crop top-1 validation accuracies for top scoring single-model architectures [18]

The training and validation accuracies over each epoch are shown in Figure 6. After 5 epochs, the model's training accuracy jumps over 90%. After 20 epochs, the training and validation accuracies attain to approximately 98%. The training and validation losses over each epoch are shown in Figure 7. The losses converge to 0.05 after 20 epochs.

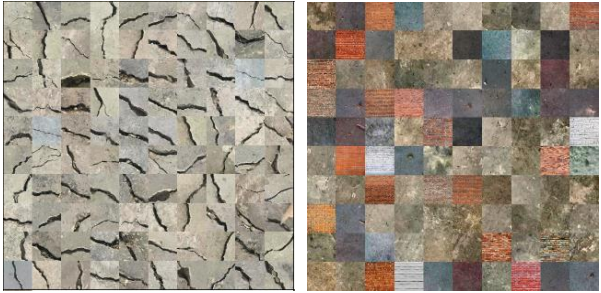

Figure 5. Sample of images used in the training ('Crack' images at left, 'NonCrack' images at right)

### 4.2 Cross-validation

After the training session, cross validation is done by using a different dataset in terms of image variation. The cross validation dataset consists of 64095 images. 19368 of these have surface cracks while there are no cracks in the rest. The fine-tuned model accurately predicts 62417 from the 64095 image. The accuracy is 97.382% in the cross validation. These results clearly demonstrate the convenience of using an InceptionV3 model as the backbone for transfer learning for such a crack detection application.

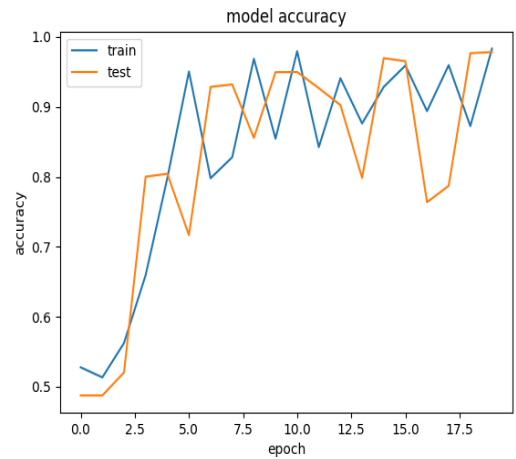

Figure 6. Accuracy vs. epoch number in the training and the validation sets

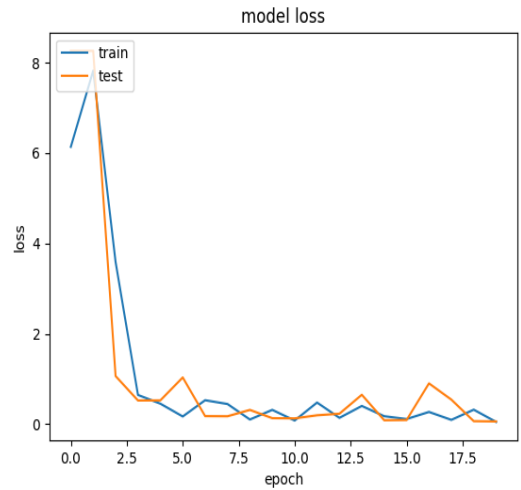

Figure 7. Loss vs. epoch number in the training and the validation sets

## 5 Experiments

In order to validate the system, autonomous navigation and crack detection capabilities are tested through computer simulations and real-world experiments.

### 5.1 Simulations

The simulations are performed in order to verify the state estimation performance of the system. For this purpose, a simulated environment is created inside Gazebo simulator. The environment is constructed to mimic an indoor space to be able to verify the performance of visual-inertial navigation of the system in GPS-denied environments. Figure 8 shows the environment used in simulations. Two connected spaces in the environment is enclosed by 3x1x3 m (height x width x length) brick walls. Another wall with relatively monotonous texture is located because identifying loop-closures is harder when repetitive patterns are present in the environment. In this way, the performance of loop closure detection is more indicative since it is a more challenging case.

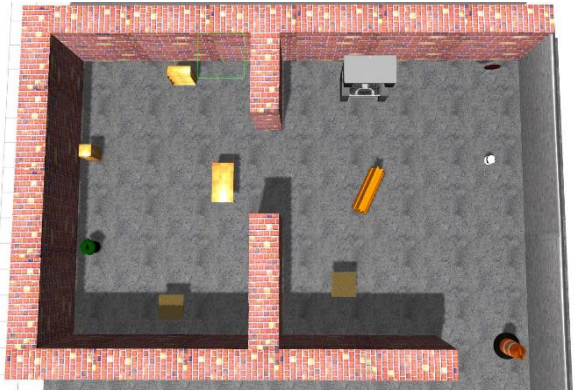

Figure 8. Indoor test environment used in the simulations

### 5.1.1 Mapping

After the simulation environment is established, the tests are performed to evaluate the performance of the mapping, the state estimation, and planning approaches. First, an exploration (of the environment) session is conducted by manually operating the quadrotor UAV, a 3D voxel grid map is constructed Figure 9. The performance of the mapping is assessed by comparing it with the original environment in the simulations.

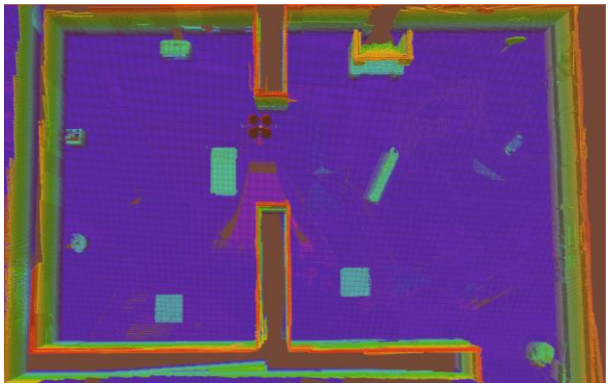

Figure 9. Reconstructed 3D voxel grid map of the environment

### 5.1.2 State Estimation

In order to evaluate the state estimation performance of the system, the position (global x-y-z) and orientation (yaw) estimates are compared with the corresponding ground truth values. The ground truth values are obtained from the simulation environment. Visual odometry and visual-inertial odometry results are plotted along with ground truth values (Figure 10). The estimates are closely tracking the actual measurements of the trajectory.

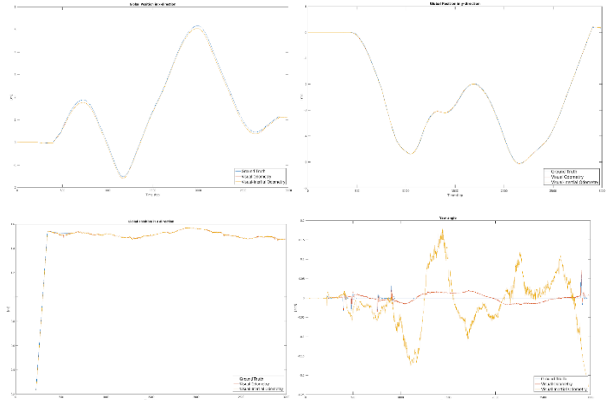

Figure 10. Ground truth vs. Visual-inertial estimations (x-direction at top-left, y-direction at top-right, z-direction at bottom-left, yaw degree at bottom-right)

The maximum deviations (errors) in state estimates are presented in Table 1 in order to comprehend the results in a clearer way. It can be observed that visual-inertial odometry has superior performance than visual odometry as expected. Although the maximum deviations in the x, y and z direction are close to each other, the errors in the yaw angle are slightly dramatic compared to them. This expected behavior shows the importance of fusing inertial measurements with visual odometry since yaw angle of IMUs are generally prone to have error due to the fact that the gravity measured by accelerometers cannot be used to help to estimate it [19]. The values in the table show that the maximum error along the estimated trajectory is in the order of 0.1 m. This value is negligible relative to the building scale so, the performance of the state estimation can be evaluated as sufficient in terms of building inspection.

Table 1. Maximum deviations in the state estimation

|  | Visual Odometry Max. Deviation | Visual-Inertial Odometry Max. Deviation |
|---|---|---|
| Global x-direction | 0.146157219 m | 0.145886557 m |
| Global y-direction | 0.064824391 m | 0.052191800 m |
| Global z-direction | 0.058057038 m | 0.054663238 m |
| Yaw Angle | 0.23723990 rad | 0.02274868 rad |

## 5.2 Indoor Experiment

The second test case is conducted indoor to be able to verify the fully integrated system. After verifying the state estimation and mapping performances in simulations, experiments are performed in a GPS-denied environment for evaluation of the integrated system. The experiments are conducted in the workshop of Design Factory in Middle East Technical University that can be seen in Figure 10.

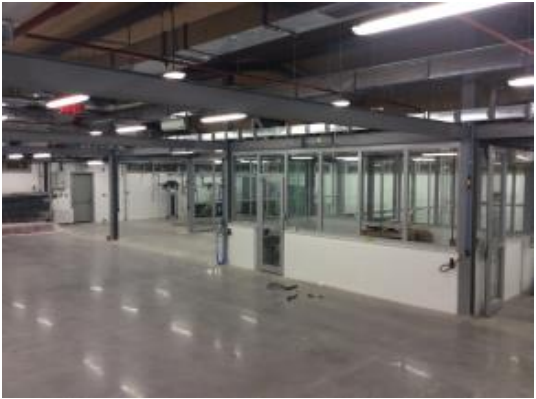

Figure 11. Indoor experimentation environment

### 5.2.1 Hardware

In real-world tests, hardware of the system is composed of five components which are an aerial platform, an onboard visual sensing system, a RGB-D camera and an onboard computer. As the aerial platform, DJI Matrice 100 [10] is employed. It is a vertical take-off and landing (VTOL) quadrotor vehicle with reconfigurable hardware installation capability. Matrice 100 is used since it meets the requirements of this research by having onboard low-level flight controller that handles attitude control.

For the onboard visual sensing system, five units of low resolution stereo camera and one processor named Guidance [20] are implemented. Guidance is compatible with the flight controller of the aerial platform, and it fuses stereo camera data for real-time obstacle avoidance in this work.

As the onboard RGB-D sensor, a widely used and open sourced hardware, Microsoft Kinect v1 is used. Having served for the purpose of mapping and localization, it has an RGB camera and infrared depth camera with 43° vertical by 57° horizontal field of view at 30 frames per second.

As the onboard computer of the system, DJI Manifold that has a quad-core, 4-plus-1 ARM processor, NVIDIA Kepler-based GeForce graphics processor, 2GB memory with customized version of Ubuntu 14.04LTS is preferred. Besides, it has a wireless connection chips and antennas for communication purposes.

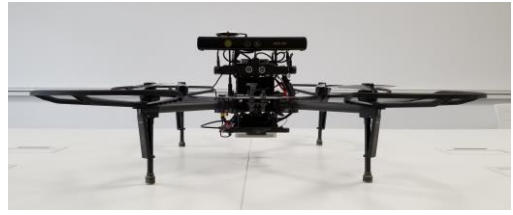

Figure 12. Integrated quadrotor UAV used in this work

### 5.2.2 Revisiting a Defected Location

The first phase of this test case is the mapping phase. A low resolution point cloud representation is presented in Figure 13. The environment is partially mapped since it is sufficient for the evaluation of the system. For mapping of the environment, the UAV is covered a trajectory (blue marker in Figure 13). The trajectory contains overlapping positions so that loop closures can be detected.

The test case is demonstration of a revisiting task. For this purpose, the images acquired during the mapping phase are processed with the image classifier developed to detect cracks on walls of the environment.

The algorithm that is developed for matching the images with their corresponding locations acquires 20 images (Figure 14) in the mapping phase. 13 of these images are replaced with images that contain cracks (Figure 15) since there are no surface cracks available in the test environment. The registered locations (positions) are kept same as in mapping but only the images are changed. In this way, the developed task planning pipeline as well as the crack detection approach can be tested.

Then, these 20 images are fed into the CNN in order to identify the cracks. After processing, the CNN classifies 16 of the images as cracks although it should be 13. Several other elements such as windows and radiator mislead the CNN as in the extra 3 images. The other 13 images are those which have cracks. Thus, it can be stated that the crack detection application gives sufficient performance as a decision support tool.

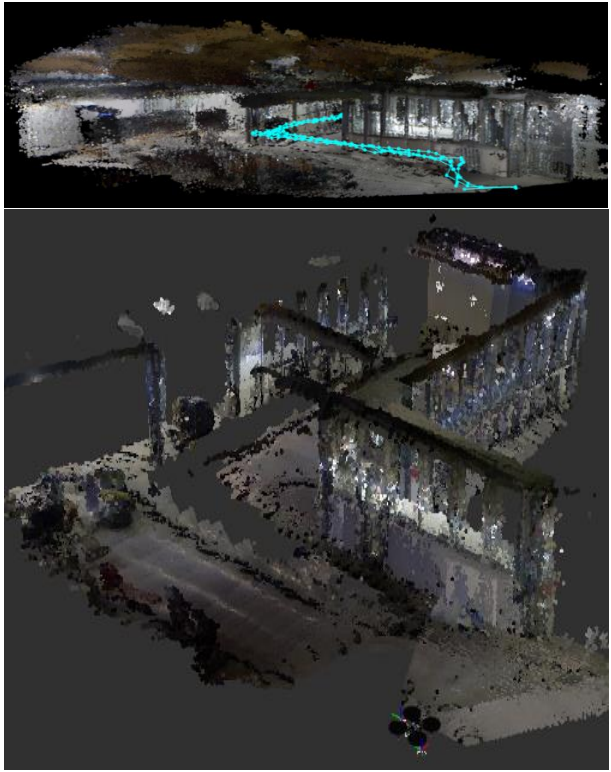

Figure 13. Reconstructed maps of the environment

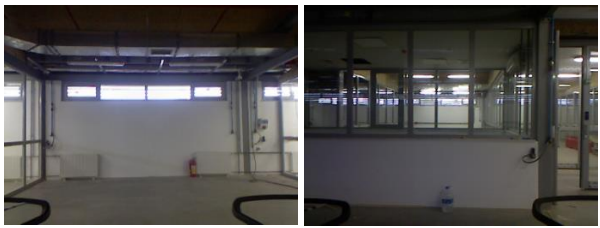

Figure 14. Sample images acquired during the mapping session

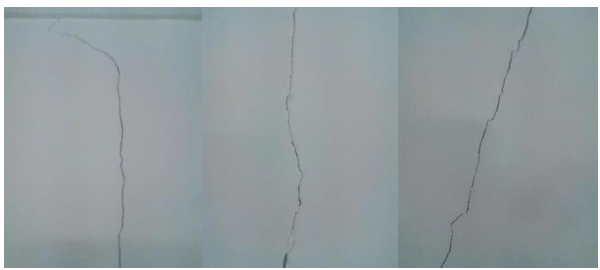

Figure 15. Sample crack images that are replaced with the acquired images

### 5.2.3 Motion Planning

After the crack detection, one of these crack images that corresponds to a specified location that is selected as the goal position for revisiting. The goal position is selected so that the most complicated motion plan should be achieved in the environment. Thus, the distance between the start and the goal positions is set to be as long as possible in the map. Moreover, the walls and the windows exist between them as obstacles so that the UAV should takeoff and move around the junction of the two walls for obstacle avoidance during motion.

Path planning algorithms that are available in MoveIt! are tested for the motion planning problem between start and goal positions. PRM*, RRT and RRT* algorithms compute solutions while EST, SPL, LBKPIECE, PRM, BKPIECE algorithms are not able to solve the problem. The reason is possibly the sampling strategies of these algorithms. They might not be able to sample the workspace such that the start and the goal positions are covered for a complete solution. The solution of the PRM* algorithm (Figure 16) is not acceptable in terms of both the optimality and the motion constraints since it requires large roll degrees in the motion that may cause overturn. The trajectory planned by RRT (Figure 16) has a sudden jump in the motion which is not possible for the UAV to execute. On the other hand, RRT* computes a trajectory (Figure 16) that satisfies criteria for motion planning. The trajectory is collision-free and smooth as well as optimal in terms of length. Therefore, the motion planner is set to use RRT* as the main algorithm in the motion plans. After the motion is planned, the UAV can be sent to the goal location for revisiting the crack location.

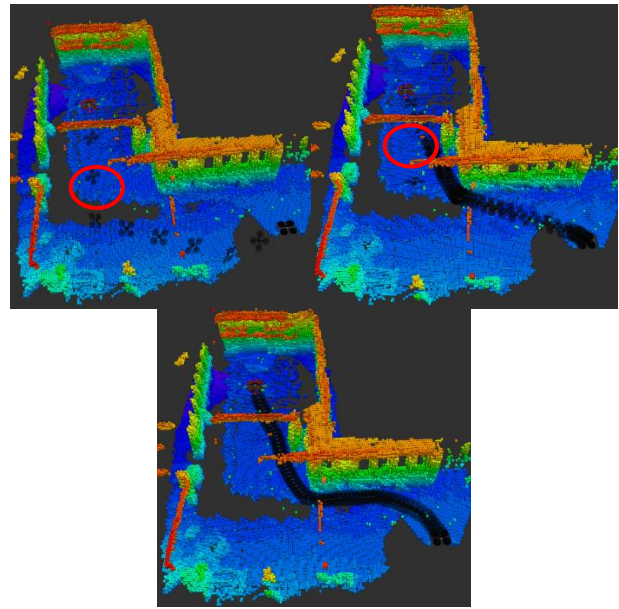

Figure 16. Motion plans (PRM* at top-left, RRT at top-right, RRT* at bottom) Red circles show the motions that violate the planning objectives

## 6  Conclusion

In this work, an integrated system that enables revisiting crack locations during building inspections by means of a quadrotor UAV is presented. Autonomous navigation of the UAV in GPS-denied environments is achieved by integrating and developing open source software. A task planning strategy is developed in order to revisit defected locations. Transfer learning is used for surface crack detection. Simulations and indoor experiments are conducted for the system verification.

The major contribution of this work can be stated as a application for building inspection by autonomous UAVs considering not only the data acquisition (mapping) phase but also the subsequent close examination (revisiting) of crack locations that are identified by a CNN. Future work will focus on improvements of crack detection on different materials and of crack properties.